\documentclass[letterpaper, 10 pt, conference]{ieeeconf}  

\IEEEoverridecommandlockouts                              

\usepackage{listings}
\usepackage{dirtytalk}
\usepackage{graphicx}
\usepackage{xcolor}
\usepackage{wrapfig}
\usepackage[colorlinks = true,
            linkcolor = blue,
            urlcolor  = blue,
            citecolor = blue,
            anchorcolor = blue]{hyperref}

\begin{document}
\title{Blending Participatory Design and Artificial Awareness for Trustworthy Autonomous Vehicles}


\author{Ana Tanevska$^{1}$, Ananthapathmanabhan Ratheesh Kumar$^{1}$, Arabinda Ghosh$^{2}$, Ernesto Casablanca$^{3}$, \\
Ginevra Castellano$^{1}$
and Sadegh Soudjani$^{2}$
\thanks{*This work is supported by the Horizon Europe EIC project SymAware (\url{https://symaware.eu}) under the grant agreement 101070802.}
\thanks{$^{1}$Ana Tanevska, Ananthapathmanabhan Ratheesh Kumar and Ginevra Castellano are with Uppsala University, Uppsala, Sweden.
        Email: {\tt\scriptsize ana.tanevska@it.uu.se, ananthapathmanabhan.ratheesh-kumar.0652@student.uu.se, ginevra.castellano.it.uu.se}
        }%
\thanks{$^{2}$Arabinda Ghosh and Sadegh Soudjani are with Max Planck Institute for Software Systems, Kaiserslautern, Germany.
        Email: {\tt\scriptsize arabinda@mpi-sws.org, sadegh@mpi-sws.org}
        }%
\thanks{$^{3}$Ernesto Casablanca is with Newcastle University, Newcastle upon Tyne, England.
        Email: {\tt\scriptsize e.casablanca2@newcastle.ac.uk}
        }%
}

\maketitle
\thispagestyle{empty}
\pagestyle{empty}

\begin{abstract}

Current robotic agents, such as autonomous vehicles (AVs) and drones, need to deal with uncertain real-world environments with appropriate situational awareness (SA), risk awareness, coordination, and decision-making. The SymAware project strives to address this issue by designing an architecture for artificial awareness in multi-agent systems, enabling safe collaboration of autonomous vehicles and drones. However, these agents will also need to interact with human users (drivers, pedestrians, drone operators), which in turn requires an understanding of how to model the human in the interaction scenario, and how to foster trust and transparency between the agent and the human. 

In this work, we aim to create a data-driven model of a human driver to be integrated into our SA architecture, grounding our research in the principles of trustworthy human-agent interaction. To collect the data necessary for creating the model, we conducted a large-scale user-centered study on human-AV interaction, in which we investigate the interaction between the AV's transparency and the users' behavior.

The contributions of this paper are twofold: First, we illustrate in detail our human-AV study and its findings, and second we present the resulting Markov chain models of the human driver computed from the study's data. Our results show that depending on the AV's transparency, the scenario's environment, and the users' demographics, we can obtain significant differences in the model's transitions.



\end{abstract}


\section{Introduction}


\say{Awareness} refers to the capacity of an agent to detect, interpret, and respond to changes in its environment, both internal and external. In autonomous agent based systems, awareness involves monitoring the operational context, identifying relevant events, and adapting behavior based on the situation. As these systems are increasingly deployed in real-world safety critical scenarios involving humans, incorporating awareness becomes essential~\cite{dignum2019responsible}. 
\begin{figure}[t]
    \centering
    \includegraphics[width=0.9\linewidth]{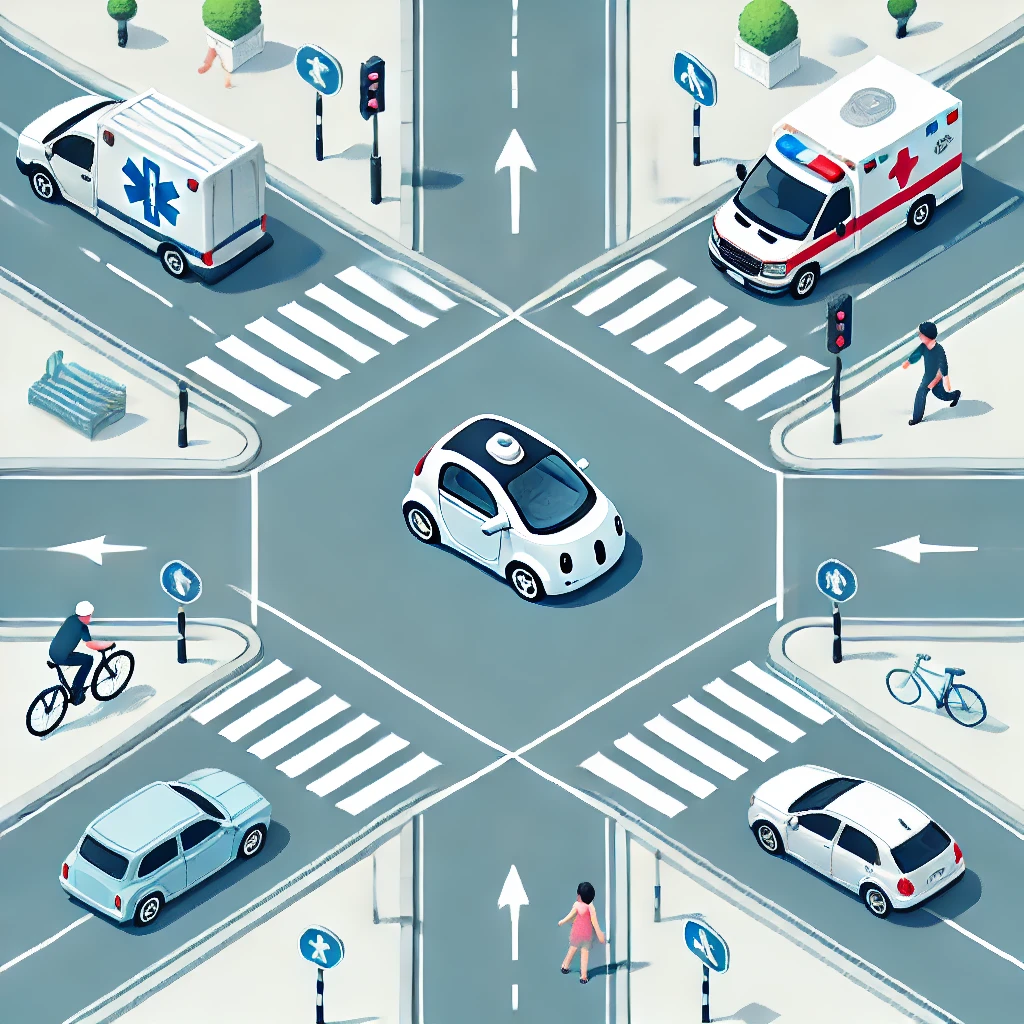}
    \caption{A complex urban intersection scenario illustrating the autonomous systems interacting with pedestrians, emergency vehicle and human-driven cars. Image generated using DALL-E.}
        \vspace{-20pt}
    \label{fig:intersection}
\end{figure}
Applications such as intelligent transportation rely on multiple agents collaborating to accomplish shared objectives. This collaborative framework is referred to as multi-agent systems (MASs), which can consist of physical robots or virtual agents~\cite{ferber1999multi}. Introducing awareness to artificial agents within a MAS scenario serves multiple objectives: awareness drives the agents to perceive and understand the intentions, actions, and states of other agents in the system; and enables them to exhibit \textit{cooperative} behavior, \textit{coordinating} their actions and collaborating towards common goals efficiently and effectively~\cite{beydoun2009security,zhang2009collaborative}. Awareness in MASs helps to develop \textit{adaptive decision-making}~\cite{endsley1995toward} and facilitate \textit{task allocation} and \textit{resource management} among the agents by efficiently \textit{planning} and \textit{executing} the tasks while \textit{resolving conflicts}~\cite{pechoucek2000coalition}.

The ability to effectively \textit{communicate} and exchange information with other agents and the environments also assists the agents in achieving their objectives. The communication process within a MAS can take various forms, such as message passing, shared memory, negotiation and agreement protocols, ontologies and semantic communication, and perception-based exchanges~\cite{stone2000multiagent}. Integrating this communication process 
can foster a sense of autonomy among the agents in the MAS (as well as between different MASs). 

While current MAS architectures focus on inter-agent awareness, an equally important challenge is enabling effective human-agent interaction. For instance, Fig.~\ref{fig:intersection} illustrates a critical urban road intersection scenario involving an autonomous vehicle, an ambulance, a regular car, a cyclist, pedestrians, and a child. In such settings, autonomous agents must not only coordinate with other vehicles and infrastructure but also interpret and respond to human behavior in real time. Effective interaction with human users, whether they are drivers, pedestrians, or cyclists, is essential for ensuring the safety of all participants, and maintaining the users' trust in the autonomous systems. 

With our research, we attempt to address precisely this issue, namely: \textit{how can artificial agents in a MAS communicate with human agents in a trustworthy manner?} We aim to bridge the gap between current research on designing and implementing MAS architectures which are centered around the artificial agents' capabilities, and the existing human-agent/human-robot interaction (HAI/HRI) research on designing trustworthy, user-centered interaction.

A first step towards building this interaction can be to incorporate a model of human agents into the MAS architecture of the artifical agent. This in turn requires an understanding of people's intentions and behaviors \cite{tanevska2020socially,meltzoff1995understanding}, which can be obtained through user-centered studies \cite{winkle2021leador}. 
Moreover, as user behavior is also affected by various aspects of the autonomous agent, we need to also consider the requirements for building a trustworthy and ethical HAI
, and investigate how the agent's autonomy will impact the human, as well as which HAI parameters will have the highest impact on the user's sense of trust in the agent \cite{khavas2020modeling,gao2019fast}.  

We begin our study design from the existing guidelines on trustworthy and ethical HAI \cite{eu_ethics_guidelines_2019}, and then proceed to explore specifically the trustworthy concepts of \textit{agent transparency} and \textit{user oversight and agency}. From these two concepts, we define the research questions to be addressed in our work:
\begin{itemize}
    \item \textbf{\textit{How can human users be in the loop and monitor the interaction?}} (which will explore the agent’s transparency and the information it shares with the human.)
    \item \textbf{\textit{To what extent should the agent act autonomously?}} (which will explore the interplay between the agent’s autonomy and the user’s preferences.)
\end{itemize}



The first interaction scenario we are working with is the one between an autonomous vehicle (AV) and a human in the role of its driver. 
Our study's design is informed from a prior participatory workshop \cite{tanevska2023communicating} on driver-AV interaction, 
where users experienced two different scenarios as an AV driver. In it, we manipulated the level of information the AV provided to the participants, and we tried to understand how the AV's transparency interacts with the users' sense of agency, as represented by the amount of autonomy users were willing to grant to the AV.

In this paper, we aim to create a data-driven model of a human driver to be integrated in our situational awareness (SA) architecture, grounding our research in the principles of trustworthy HAI. To collect the data necessary for creating the model, we conduct a large-scale user-centered study on human-AV interaction, in which we investigate the interaction between the AV's transparency and the users' behavior.
Our contributions with this research are twofold: we illustrate in detail our human-AV study and its findings, and we present the resulting Markov chain model of the human driver computed from the study's data.


The rest of the paper is organized as follows: Section ~\ref{section:Architecture} presents our MAS architecture for situational awareness, Section ~\ref{section:Methodology} presents our human-AV study on transparency and user behavior, Section ~\ref{section:model} presents the data analysis and building of the user model, and finally Section ~\ref{section:conclusion} summarizes our findings and outlines our next steps.

\section{MAS Architecture for Situational Awareness}\label{section:Architecture}

\begin{figure*}[t]
    \centering
    \includegraphics[width=0.8\textwidth]{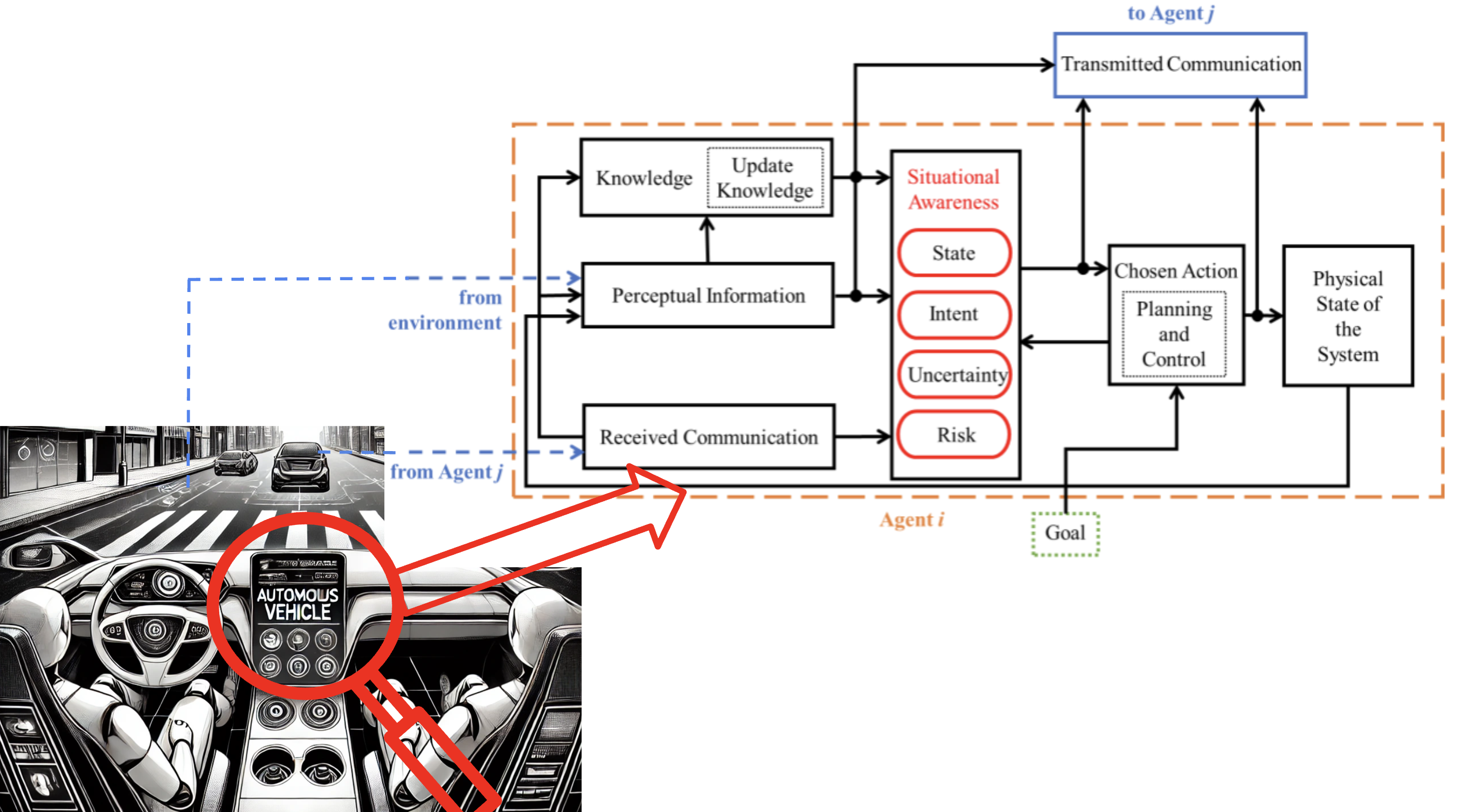}
    \vspace{-10pt}
    \caption{The SymAware architecture provides insights into the internal structure of an agent $i$, which interacts with another agent $j$ through the communication module.}
    \label{fig:architecture}
\end{figure*}

Designing effective MASs requires a structured architectural approach that enables agents to operate autonomously while maintaining coherence in complex, dynamic environments. One critical aspect of MAS design is situational awareness (SA), which allows agents to perceive, interpret, and respond to their surroundings in a way that ensures robustness and reliability. Situational awareness in MASs extends beyond individual agents as it emerges from distributed sensing, knowledge integration, cooperative reasoning, and interactions with human entities, ensuring that the system as a whole can anticipate and adapt to changes, uncertainties, and potential risks. By embedding SA-driven decision-making within the architectural framework, MASs can achieve higher levels of resilience, autonomy, and safety, particularly in safety-critical and interactive environments.   

This is where SymAware becomes particularly useful, providing a structured framework for developing MAS architectures with advanced SA. SymAware proposes a novel cognitive architecture, founded on compositional logic, symbolic computations, formal reasoning, and uncertainty quantification, enabling agents to dynamically assess their surroundings, quantify risks, and adapt to changing conditions. The framework introduces logical characterizations of awareness, spatiotemporal reasoning for task planning, and risk quantification, facilitating robust coordination and decision-making. Unlike conventional MAS architectures, SymAware explicitly incorporates knowledge awareness and risk-aware negotiation, allowing agents to interpret limited information, anticipate risks, and reason about uncertainties. By embedding these capabilities, SymAware ensures that autonomous agents operate in a verifiable, explainable, and ethically aligned manner, supporting safe interactions between AI-driven agents and humans. Our MAS architecture as shown in Fig.~\ref{fig:architecture}, initially introduced in our previous work~\cite{tanevska2023communicating}, provides the foundation for these capabilities. Below we will briefly outline its key components:

\textbf{\textit{- Knowledge.}} In a MAS, the knowledge base (KB) acts as a structured repository where an agent stores, organizes, and retrieves information necessary for coordination, learning, adaptation, and decision-making. In the SymAware architecture, the KB functions as a repository where an agent stores and manages information about the environment, particularly information related to traffic signs and regulations. This stored knowledge enables agents to interpret road regulations, navigate effectively, and make decisions aligned with predefined traffic rules. By integrating perceptual inputs and refining stored knowledge over time, the system ensures that agents can respond accurately to their surroundings and operate safely within the MAS.

\textbf{\textit{- Estimation and Perception.}} The estimation and perception component enables agents to interpret their surroundings and extract meaningful information from the environment. This block processes data collected through various sensing mechanisms, such as cameras, LiDAR in order to recognize traffic signs and other relevant environmental features. By applying object recognition, and intelligent algorithms, raw sensory inputs are transformed into a structured representation that the agent can use for decision-making and SA component. This ensures that agents operate with an up-to-date and reliable perception of their environment, supporting safe and efficient navigation.

\textbf{\textit{- Communication.}} Communication module enables agents to exchange information and signals, facilitating coordination, collaboration, and adaptation within the MAS. By sharing relevant data, agents enhance collective awareness and improve decision-making, ensuring more effective and intelligent system behavior.

\textbf{\textit{- Situational Awareness (SA).}} In the SymAware architecture, situational awareness enables agents to perceive changes, predict potential risks, and adapt their actions accordingly. The SA module consists of four key aspects: state awareness, which integrates the perception of the agent with environmental factors such as obstacles and road elements; intent awareness, which allows the agent to infer the goals and behaviors of other entities; uncertainty management, ensuring robustness in decision-making despite incomplete or ambiguous information; and risk assessment, which helps detect hazards and anticipate threats.

\textbf{\textit{- Planning and Control.}} In the architecture, this module translates SA into concrete actions. The planning component formulates strategies aligned with the objectives of the agent, while the control component executes these strategies, ensuring the system operates as intended.

\textbf{\textit{- Physical State of the System.}} The physical state represents the dynamic behavior of the agent, which evolves based on the control actions executed. These changes, along with environmental updates, are fed back into the estimation and perception module that ensures adaptability and informed decision-making.


\section{Methodology}\label{section:Methodology}
In order to bridge the gap between designing agent-centered architectures, and designing trustworthy human-agent and human-robot interaction (HAI/HRI), we worked on creating a representation of the human user to be integrated in our MAS architecture. This representation needed to be computed from real human data, which we sought to obtain from people in a participatory, user-centered manner. Conducting participatory studies with the potential end users allowed us to uncover any ethical concerns users may have, as well as explore in-depth the trustworthy concepts we wish to represent.

In our research, the two main trustworthy concepts we wished to explore were \textit{transparency} and \textit{agency}, which we mapped in the following research questions:

\begin{itemize}
    \item \textit{How can human users be in the loop and monitor the interaction?} - which explored the agent's transparency as represented by the amount of information it shares with the human.
    \item \textit{To what extent should the agent act autonomously?} - which explored the interplay between the agent's autonomy and the user's preferences. 
\end{itemize}

We began by conducting a small participatory design study \cite{tanevska2023communicating} consisting of two different driver-AV interaction scenarios, where we worked together with the participants to develop different interfaces for the AV for their preferred levels of information sharing (i.e. \textit{transparency}), as well as analyzed how comfortable they were with the different levels of autonomy (LoA) in which the AV functioned.

We then followed the participatory workshop with a larger-scale online study, whose design was informed by the findings of our workshop. With it, we manipulated the level of information the AV provided to the participants, and we tried to understand how the AV's transparency interacts with the users' sense of agency. This study's purpose was to collect the data for creating the model of the human driver, and we sought to understand how the different transparency levels of the AV and the different interaction scenarios impact the level of autonomy the users would grant to the AV. 

\subsection{Study Background}
\label{sub:background}
As a first step in our study's design, we conducted a small participatory design workshop \cite{tanevska2023communicating}, the purpose of which was to understand the preference of human users' with respect to the AV's autonomy in different scenarios, as well as the AV's interface and the amount of information it can communicate.

\begin{figure}[t]
    \centering
    \includegraphics[width=0.75\columnwidth]{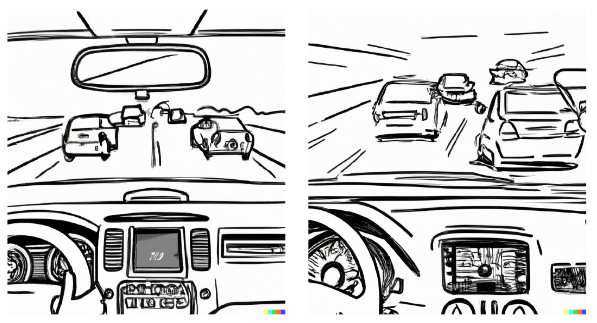}
    \caption{\textit{Highway} driver-AV scenario. AI-generated images of a car's dashboard with the driver's POV, car driving on a busy highway and another car cutting in from the right.}
    \label{fig:scenario1}
    \includegraphics[width=0.75\columnwidth]{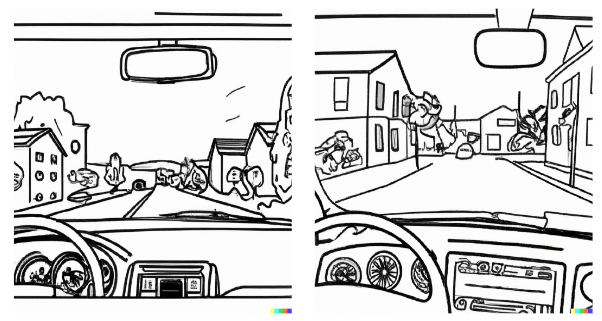}
    \caption{\textit{Suburbs} driver-AV scenario. AI-generated images of a car's dashboard with the driver's POV, car driving on an empty neighborhood street, with a school building coming up on the left.}
    \label{fig:scenario2}
\end{figure}

Participants worked in two use case scenarios\footnote{Presented to users as series of computer-generated images from https://openai.com/dall-e-2 on a flip-chart}, where they took on the role of the AV's driver. In the first scenario (Fig. \ref{fig:scenario1}), the AV was driving along the middle lane on a busy highway, when another car unpredictably merges from the right at an unsafe distance and speed, creating an imminent collision risk. 
In the second scenario (Fig. \ref{fig:scenario2}), the AV was driving down an unfamiliar empty suburban street, when it suddenly slowed down to 20 km/h without any visible obstacle. As the scenario progressed, a school that was previously obscured behind trees came into view.

\begin{figure}[h]
    \centering
    \includegraphics[width=0.9\columnwidth]{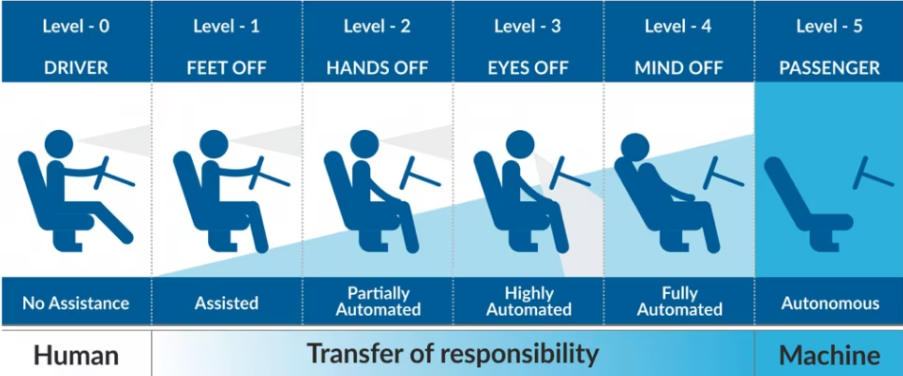}
    \caption{The six SAE Levels of Autonomy. In our study we worked with Levels 0 to 3, i.e. "Driver", "Feet off", "Hands off", and "Eyes off".}
    \label{fig:sae}
\end{figure}

The workshop had two activities - modulating the \textbf{Level of Autonomy (LoA)}, and designing a \textbf{user interface} for interaction between the car and the driver. The levels of autonomy we worked with in this study were the SAE Levels of Autonomy \cite{sae2018taxonomy}, specifically Levels 0 through 3 (see Fig. \ref{fig:sae}). For each scene, participants were asked to select the highest level of autonomy they were comfortable with.

We then presented participants with the information the AV uses to calculate its SA and select its next action, and asked them to design an interface for the AV (using post-it notes and drawing on the flip-chart), deciding on the amount and type of information shown. The workshop produced two distinct interface designs: 1. A minimal, unobtrusive interface similar to modern car GPS navigators; and 2. A much more complex design containing visual depiction of all of the information the AV is processing and its upcoming actions.

The two distinct interfaces thus informed the design of our large-scale online study (presented in the rest of this section) and became the two information levels - with the minimal interface giving us the Low Information condition, and the complete interface the High Information one.



\subsection{Participants}
\label{sub:participants}
In total 206 participants took part in our study, equally distributed between male and female\footnote{Legal sex as reported on their current documents. The self-reported gender ratio instead was 104:101:1 (F:M:X)}, and their age ranged from 20 to 73 years (M=36.42, SD=13.185). Participants were recruited via the crowd-sourcing platform Prolific\footnote{https://www.prolific.com/}. We used the platform's filters to recruit English-speaking participants with a valid driver's license. 


\subsection{Experimental Design and Protocol}
\label{sub:experiment}
Our study followed a 2×2 between-subjects design, corresponding to the designed \textit{information level} provided by the AV's interface (High Information vs. Low Information) and the \textit{order} in which participants experienced the two scenarios (Highway-first vs. Suburbs-first). Participants were randomly assigned to one of the four groups.

At the start of the study, participants completed questionnaires on demographics and driver behavior (an adapted, abridged version of the DBQ \cite{reason1990errors}). 
This was followed by a brief presentation on AVs which introduced participants to the term, outlined the types of information the AV processes when making its decisions, and presented participants with the concepts on autonomy in AVs and how it's quantified (here we also used the SAE Levels of Autonomy, see Fig. \ref{fig:sae}). It also broached the concept of social interaction with AVs.

After the presentation, there was a manipulation check (describing one SAE level and asking participants to select the correct one), after which participants filled the last questionnaire, an adaptation of the NARS \cite{nomura2004psychology}, aimed at identifying and quantifying any negative attitudes or preconceptions participants may have towards AVs. Our version (referred to as AV-NARS) had 10 items, adapted to fit the context of interaction with AVs, e.g. \say{I would feel nervous driving an autonomous car while other people could observe me.} instead of the original \say{I would feel nervous operating a robot in front of other people.}. 

After the questionnaires, participants were presented with the same two scenarios (Highway and Suburbs) from the participatory design workshop as described in Sect. \ref{sub:background}. Both scenarios consisted of three scenes, with the second scene being the one with the unexpected moment (the second car merging unsafely, or the AV inexplicably slowing down).

\begin{figure}[t]
    \centering
    \includegraphics[width=\columnwidth]{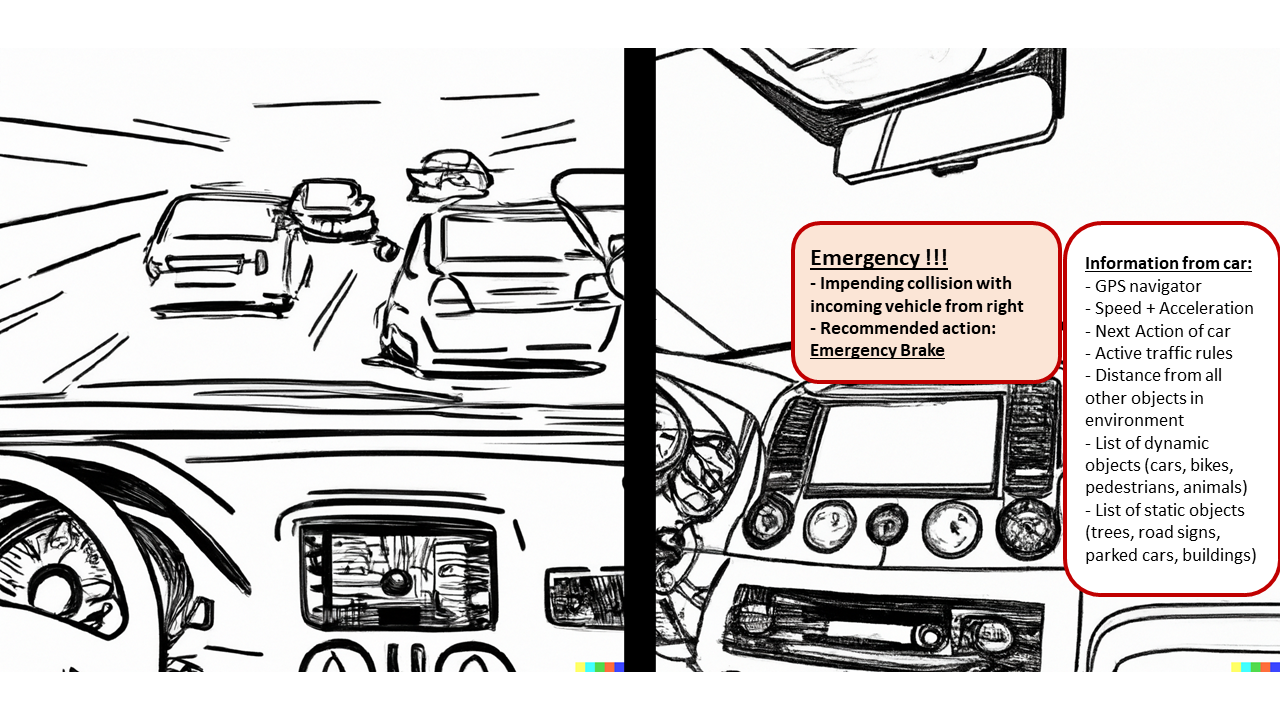}
    \caption{\textit{Highway} driver-AV scenario, plus a High Information interface for the AV. AI-generated images of a car's dashboard with the driver's POV, car driving on a busy highway and another car cutting in from the right.}
    \label{fig:scenarioFull}
\end{figure}

Each of the three scenes consisted of a computer-generated image of the dashboard view of the scene plus an overview of the information presented to them (see Fig. \ref{fig:scenarioFull}), and a roleplay-like narration of what happens in the scene. Participants were then asked "What do you do?" and could select from multiple-choice answers, where each choice corresponding to a level of autonomy between 0 and 3. E.g. for the first scene of Scenario 1, the possible answers were:
\begin{itemize}
    \item I do nothing, I let the car continue navigating
    \item I quickly check my phone for messages
    \item I check out the area around me and look around for anything interesting
    \item I focus my eyes on the road and check for any people or animals that may approach the roa
    \item I put my hands back on the wheel
    \item I take over the driving control from the car
\end{itemize}

After selecting an action, participants answered questions on their confidence and comfort levels with the situation, and their trust in the AV. The questions are described in greater detail in the following section.

\subsection{Data Collection}

We grouped the collected data in four categories:

\emph{- Questionnaires.}
Participants answered questionnaires on demographics, adapted DBQ (driver behavior questionnaire) and adapted AV-NARS (negative attitudes towards AVs).

\emph{- Action choice (Level of Autonomy).}
The selection of action following each of the six scenes. A multiple-choice question where each choice corresponded to a level of autonomy between 0 and 3. This was averaged in the cases where participants chose more than one option, with the final LoA score thus being a decimal value ranging from 0 to 3.

\emph{- Confidence and comfort.}
After each action decision, participants were asked to rate how confident they felt about their decision (5-point Likert scale with descriptive ranking, ranging from \textit{\say{Not confident at all}} to \textit{\say{Completely confident}}). They were also asked about their comfort with the situation, which was a single-choice question where they could pick \textit{\say{relaxed, calm}} (rated 1), \textit{\say{anxious, tense}} (rated -1), \textit{\say{neutral}}, and \textit{\say{unsure}} (both rated 0).

\emph{- Trust in the AV.}
For their final evaluation, participants were asked about their trust in the car via a multiple choice question, where they could rate the car as reliable, safe, and trustworthy.
The answers were coded as 1 or -1 depending on whether they expressed a positive or negative trust sentiment, and then summed up to form the total Trust score, which ranged from -2 to 4. Participants could also provide a free-text answer, if they felt that none of the statements were accurate about the current situation and their decision.







\section{Creating a model of the human driver}\label{section:model}
The second goal in our research was to create a computational model of the human driver based on the data on their behaviour we obtained from our user study.
Specifically, we wished to see how the autonomy of the AV was perceived by the users, and how they interacted with it throughout the different scenes in the different scenarios. 

We wanted to understand if the Level of Autonomy (LoA) selected by the users depended only on the scenarios and external factors in each scene, or if it was also influenced by other elements, such as the level of information of the AV, or the participants' sex. 
This section presents the work we did on building a mathematical model of the driver data, and the analysis of how users interacted with the AVs over time.

\subsection{Modeling human behaviour with Markov Chains}

Markov chains are mathematical models with finite number of states. The Markov chain will jump from one state to another following certain probabilities. The hallmark feature of a Markov chain is that the succeeding state depends solely on the current state instead of how the system arrived at the current state \cite{norris1997jr}. 

This \say{memory-less} feature makes Markov chains efficient for designing systems where the future state depends only on the present state. Such models can also be used for analyzing systems that have finite memory by extending the state space of the model. Utilizing Markov chains to study properties of systems have been studied extensively in the literature \cite{KNP11,SA13}. 
Moreover, Markov models have also been applied for describing and categorizing human behavior at various levels of abstraction \cite{theocharous2002hierarchical,pentland1999modeling}. 

In our study, we used Markov chains to design the model of the human driver behavior \cite{camara2020pedestrian}, and specifically model how the users' actions (their preferred level of autonomy) change depending on the scenario's progress and other factors.
We opted to begin with simple Markov chains as a way to better understand the user data and see how well the model describes it, keeping open the possibility of later upgrading the model to a more complex one, such as Hierarchical or Abstract Hidden Markov Models \cite{fine1998hierarchical,bui2002policy}.


\subsection{Building a Markov Chain from Data}

To create our Markov chains, we began by identifying and defining the model's states. In addition to the model's Start state, we defined the following three states: Takeover (T), Alert (A), and Normal (N). The states corresponded to intervals of the LoA value, specifically:
\begin{itemize}
    \item T (Takeover): \(LoA <= 0.5\)
    \item A (Alert): \(0.5 < LoA < 2.0\)
    \item N (Normal): \(LoA >= 2.0\)
\end{itemize}

The Start state was by default set as Normal, since at the beginning of each scenario users were told the AV was operating at a LoA of 3.
After mapping the LoA value from each scene in both scenarios to a corresponding model state, we then proceeded to analyze the distribution of states in Scenes 1, 2, and 3.
Tables \ref{tab:initial-highway} and \ref{tab:initial-suburbs} show the initial distribution of states for Scene 1 of both scenarios:

\begin{table}[h]
\centering
\caption{Transition to Highway Scene 1 from the Start State}
\label{tab:initial-highway}
\begin{tabular}{lc}
\hline
Category & Count \\
\hline
N        & 63.5 \%   \\
A        & 27.8 \%    \\
T        & 8.7 \%    \\
\hline
\end{tabular}
\end{table}

\begin{table}[h]
\centering
\caption{Transition to Suburbs Scene 1 from the Start State}
\label{tab:initial-suburbs}
\begin{tabular}{lc}
\hline
Category & Count \\
\hline
N        & 89.8 \%   \\
A        & 8.3 \%   \\
T        & 1.9 \%    \\
\hline
\end{tabular}
\end{table}

The transition probabilities between Scenes were calculated by computing the percentage of users that transitioned from each state to each state. For instance, to calculate the probability of transitioning from Alert to Takeover between Scene 1 and 2 in the Highway scenario, we considered the 57 users who were in Alert in Scene 1, out of which 25 transitioned to Takeover in Scene 2, so the probability was \(25/57=43.9\%\).
Tables \ref{tab:highway-markov-1-2} and \ref{tab:highway-markov-2-3} show the Highway transition probabilities for Scene 1→2 and 2→3 respectively, and Tables \ref{tab:suburban-markov-1-2} and \ref{tab:suburban-markov-2-3} show the same probabilities for the Suburban scenario.

\begin{table}[ht]
\centering
\caption{Markov Transition Matrix for Highway (Scene 1→2)}
\label{tab:highway-markov-1-2}
\begin{tabular}{lccc}
\hline
Initial State & To Alert & To Normal & To Takeover \\
\hline
Alert         & 52.6\%   & 3.5\%     & 43.9\%      \\
Normal        & 37.4\%   & 35.1\%    & 27.5\%      \\
Takeover      & 16.7\%   & 0.0\%     & 83.3\%      \\
\hline
\end{tabular}
\end{table}

\begin{table}[ht]
\centering
\caption{Markov Transition Matrix for Highway (Scene 2→3)}
\label{tab:highway-markov-2-3}
\begin{tabular}{lccc}
\hline
Initial State & To Alert & To Normal & To Takeover \\
\hline
Alert         & 47.6\%   & 20.7\%    & 31.7\%      \\
Normal        & 35.4\%   & 29.2\%    & 35.4\%      \\
Takeover      & 14.5\%   & 9.2\%     & 76.3\%      \\
\hline
\end{tabular}
\end{table}

\begin{table}[t]
\centering
\caption{Markov Transition Matrix for Suburban (Scene 1→2)}
\label{tab:suburban-markov-1-2}
\begin{tabular}{lccc}
\hline
Initial State & To Alert & To Normal & To Takeover \\
\hline
Alert         & 35.3\%   & 11.8\%    & 52.9\%      \\
Normal        & 25.4\%   & 68.1\%    & 6.5\%       \\
Takeover      & 25.0\%   & 0.0\%     & 75.0\%      \\
\hline
\end{tabular}
\end{table}

\begin{table}[h]
\centering
\caption{Markov Transition Matrix for Suburban (Scene 2→3)}
\label{tab:suburban-markov-2-3}
\begin{tabular}{lccc}
\hline
Initial State & To Alert & To Normal & To Takeover \\
\hline
Alert         & 37.0\%   & 48.1\%    & 14.8\%      \\
Normal        & 12.5\%   & 80.5\%    & 7.0\%       \\
Takeover      & 25.0\%   & 25.0\%    & 50.0\%      \\
\hline
\end{tabular}
\end{table}

In addition to computing the separate Markov chains for the two different environments (highway scenario vs. suburbs scenario), we also computed the state distributions across the different information levels (high vs. low) and participant sex (male vs. female).

\subsection{Analysis and Results}

After computing all individual Markov chain models, we conducted statistical analyses to determine if the transition probabilities varied notably across conditions. We carried out chi-square tests to compare the transition patterns between environment, information level, and sex. 

\subsubsection{Environment Effect on Markov Chains} The comparison between the highway and suburban environment showed statistically significant variations in the transition patterns ($\chi^2  = 39.76,~p < 0.001$). This validates that the environment has a substantial effect on the transition of drivers between the states.

\subsubsection{Information Level Effect on Markov Chains} When comparing information levels, we found that the transitioning patterns in the highway environment showed some variations. While we observed some strong differences in the highway and suburbs transitions between certain states in certain scenes, this effect however was not present across all transition matrices, and there was no statistically significant difference at the conventional threshold ($\chi^2 = 2.24,~p = 0.326$). In the suburban environment, the information level had an even less effect on transitions ($\chi^2 = 1.04,~p = 0.593$). Table \ref{tab:info-highway-markov} shows a comparison of some key transition probabilities for the high and low information conditions in the Highway scenario.
\begin{table}[ht]
\centering
\caption{Information Level Effect on Selected Highway Transitions (Scene 1→2)}
\label{tab:info-highway-markov}
\begin{tabular}{lcc}
\hline
Transition & High Information & Low Information \\
\hline
Alert → Takeover     & 33.3\%         & 51.5\%         \\
Normal → Takeover    & 34.3\%         & 19.7\%         \\
Takeover → Takeover  & 87.5\%         & 80.0\%         \\
\hline
\end{tabular}
\end{table}

\subsubsection{Sex Effect on Markov Chains} Sex analysis revealed statistically significant variations in transition patterns for both highway ($\chi^2 = 6.13,~p = 0.047$) and suburban ($\chi^2 = 6.03,~p = 0.049$) environments, thus validating that male and female participants tend to interact differently with self-driving systems. Table \ref{tab:gender-highway-markov} shows key transition probabilities for men and women in the highway environment.

\begin{table}[ht]
\centering
\caption{Sex Effect on Selected Highway Transitions (Scene 1→2)}
\label{tab:gender-highway-markov}
\begin{tabular}{lcc}
\hline
Transition & Female   & Male     \\
\hline
Alert → Takeover     & 43.8\%  & 44.0\%  \\
Normal → Takeover    & 34.5\%  & 21.9\%  \\
Takeover → Takeover  & 76.9\%  & 100.0\% \\
\hline
\end{tabular}
\end{table}

\subsection{Statistical Validation of Markov Models}

To further validate our Markov chain models, we carried out additional statistical tests on the transition matrices.

1. \textbf{Stationarity Test}: We tested whether the transition probabilities remained consistent between scene 1→ 2 and scene 2→3. For both the highway ($\chi^2= 8.92, ~p = 0.030$) and the suburbs ($\chi^2 = 10.47, ~p = 0.015$) environment, we found notable variations, showcasing the non-stationary behavior as drivers advanced through scenarios.

2. \textbf{Homogeneity Test}: We checked if the different subgroups (divided by sex and information level) have the same underlying transition matrix. The test revealed notable differences between the sex subgroups ($\chi^2 = 9.84, ~p = 0.043$) but not between the information level subgroups ($\chi^2 = 7.31, ~p = 0.120$). 

3. \textbf{Goodness-of-Fit Test}: Finally, we assessed how well our first-order Markov model (which considers the current state only) fits into the observed data compared to a second-order model (which considers the previous two states). The likelihood ratio
test revealed that the second-order model provided a significantly better fit for the highway data ($G^2 = 14.29, ~p = 0.027$) but not for suburban data ($G^2 = 5.83,~p = 0.442$).

These statistical tests not only validate our Markov models but also highlight the important nuances in driver behavior transitions across different conditions.

\subsection{Discussion of Markov Chain Analysis}

With our statistically validated Markov chain analysis we obtained several key insights about driver behavior:
\begin{enumerate}
    \item \textbf{State Stability}: The Takeover state was the most stable across both environments, with $75-100$\% probability of the driver remaining in Takeover once entered. This suggests that users were highly unlikely to give back control to the AV once they reach the point in the interaction where they wish to take over. 
    
Additionally, in the suburban environment, the Normal state was also highly stable, with $68-80$\% probability of the driver remaining in it. 
    \item \textbf{Environmental Differences}: Highway driving showed more dynamic transitions between states and lesser tendency for the drivers to leave the AV in a high Level of Autonomy, as seen in the lower stability in the Normal state ($29-35$\% probability of remaining in Normal) compared to suburban driving ($68-80$\%).
    
This aligns with our preliminary findings that users were overall more likely to give a higher LoA to the AV in the Suburbs scenario.
    \item \textbf{Higher-Order Effects}: Our statistical analysis revealed that in highway environment, a driver’s history (not just their current state) significantly influenced their next state ($p = 0.027$). This suggests that more complex models may be needed for fully capturing highway driving behavior, aligning with our point at the beginning of this section.

\end{enumerate}




\section{Conclusion}\label{section:conclusion}


In this paper, we presented our work towards creating a data-driven model of a human driver and implementing it in the SymAware MAS architecture. Our contributions consist of: 1) designing and conducting a large-scale user-centered study on human-AV interaction, exploring the concepts of the AV's transparency and autonomy, 2) analyzing the obtained user data to extract information on the user behavior and how it developed across the scenarios, and 3) using these findings to create a computational model of the human driver, represented as series of Markov chains.

With these statistically validated Markov models we lay out a mathematical framework for understanding and predicting driver behavior in self-driving systems. The notable differences we found between different environments (\( p < 0.001 \)) and participant's sex (\( p < 0.05 \)) highlight the importance of developing adaptive systems that can respond differently based on these factors. Such systems could foresee driver state changes and provide appropriate support at the right time, potentially improving both safety and user experience. Thus, by understanding and modeling driver behavior, we take a step toward autonomous agents that are not only situationally aware but also ethically and socially aware in their interactions with humans.

Having created the initial computational model of the human driver, our next steps will be to program and verify the model using the PRISM tool \cite{KNP11}, and then integrate the verification procedure and its adaptation into our MAS architecture \cite{calvagna2023using}. We view this integration as having the potential to improve the overall performance and effectiveness of the MAS by leveraging the cognitive capabilities of both humans and machines, and to contribute to more natural, trustworthy, and long-term interactions between humans and artificial agents.

\bibliographystyle{IEEEtran.bst}
\bibliography{IEEEfull}

\end{document}